%% file: main.tex
\documentclass[fleqn,10pt]{wlscirep}
\usepackage[utf8]{inputenc}
\usepackage[T1]{fontenc}
\usepackage{amsmath}

\usepackage{makecell}
\usepackage{geometry}
\usepackage{changepage}
\usepackage{multirow}

\usepackage{booktabs,siunitx,array,threeparttable}
\usepackage{xcolor,colortbl}
\definecolor{Gray}{gray}{0.85}
\usepackage{subfigure}

\usepackage{cleveref}

\title{Measuring Transparency in Intelligent Robots}

\author[1,*,+]{Georgios Angelopoulos}
\author[2,**,+]{Dimitri Lacroix}
\author[2]{Ricarda Wullenkord}
\author[1]{Alessandra Rossi}
\author[1]{Silvia Rossi}
\author[2]{Friederike Eyssel}
\affil[1]{ Interdepartmental Center for Advances in Robotic Surgery - ICAROS, University of Naples Federico II, Naples, 80131, Italy}
\affil[2]{Center for Cognitive Interaction Technology - CITEC, Bielefeld University, Bielefeld, 33619, Germany}

\affil[*]{georgios.angelopoulos@unina.it}
\affil[**]{dimitri.lacroix@uni-bielefeld.de}
\affil[+]{these authors contributed equally to this work}


\begin{abstract}
As robots become increasingly integrated into our daily lives, the need to make them transparent has never been more critical. Yet, despite its importance in human-robot interaction, a standardized measure of robot transparency has been missing until now. This paper addresses this gap by presenting the first comprehensive scale to measure perceived transparency in robotic systems, available in English, German, and Italian languages. Our approach conceptualizes transparency as a multidimensional construct, encompassing explainability, legibility, predictability, and meta-understanding. The proposed scale was a product of a rigorous three-stage process involving 1,223 participants. Firstly, we generated the items of our scale, secondly,  we conducted an exploratory factor analysis, and thirdly, a confirmatory factor analysis served to validate the factor structure of the newly developed TOROS scale. The final scale encompasses 26 items and comprises three factors: \textit{Illegibility}, \textit{Explainability}, and \textit{Predictability}. TOROS demonstrates high cross-linguistic reliability, inter-factor correlation, model fit, internal consistency, and convergent validity across the three cross-national samples. This empirically validated tool enables the assessment of robot transparency and contributes to the theoretical understanding of this complex construct. By offering a standardized measure, we facilitate consistent and comparable research in human-robot interaction in which TOROS can serve as a benchmark.
\end{abstract}
\begin{document}

\flushbottom
\maketitle
%
%
\thispagestyle{empty}


\section*{Introduction}

\input{1_introduction}

\section*{Stage 1}

\input{2_item_generation}

\section*{Stage 2}

\input{3_experiment1}
\section*{Stage 3}

\input{4_experiment2}

\section*{Discussion}

\input{5_discussion}

\section*{Conclusions}

\input{6_conclusions}

\section*{Data availability}
The datasets generated and analyzed during the current study are available from the corresponding authors upon request.

\bibliography{sample}

\section*{Acknowledgements}
The authors thank Francesco Vigni and Lena Schubert for translating the scale into Italian and German language, respectively. This work has been supported by the European Union's Horizon 2020 research and innovation programme under the Marie Sk\l{}odowska-Curie grant agreement No 955778.

\section*{Author contributions statement}

G.A., D.L., and F.E. developed the scale. G.A. and D.L. designed the experiments.  R.W., A.R., S.R., and F.E. gave feedback on and supported the development of the scale and experimental material. G.A. and D.L. conducted the experiments and analyzed the results. R.W. translated the scale into German. A.R. translated the scale into Italian. G.A. and D.L. wrote the initial draft.  R.W., A.R., S.R. and F.E. gave feedback on and refined subsequent drafts of the manuscript. All co-authors have read and agreed to the submitted version of the manuscript.

\section*{Competing interests}
The authors declare no competing interests.

\end{document}

%% file: 1_introduction.tex

We want to invite you to imagine that you are in your kitchen with a friend, and you would like to get an ice cream. You have some in your freezer, but how do you convey your intention to get it to your friend? A possibility would be by explicitly stating ``I want to get some ice cream!''. You could also alternatively use non-verbal signals such as walking towards, gazing at, or pointing at the freezer \cite{wubben_how_2009, canigueral_role_2019}.
Verbal and non-verbal behaviors that signal intentions enable mutual understanding within human social interactions. Thereby, ambiguities and potential misunderstandings can be reduced, and accordingly, a human's behavior may be deemed transparent \cite{bao_research_2016, ahmad_review_2022}. 

Similarly, transparency is essential for human-robot interaction (HRI). Indeed, robots are designed to eventually co-exist with humans in various contexts, including the home, work, or school. Their role is to assist with daily tasks, improve the efficiency of workflows, and enhance the quality of life of individuals \cite{vigni2023sweet}. For instance, a service robot might help with household chores, assist in medical procedures, or facilitate collaborative work \cite{bhatti2024working}. For autonomous robots to function effectively and harmoniously alongside humans, it is essential that they are capable of communicating their intentions transparently, just as humans do. Previous studies have highlighted the importance of transparency for HRI as it fosters trust and acceptance between humans and robots \cite{schor2024mind,fischer2018increasing,aquilino2024trust}. Transparency helps to regulate end-users' expectations and enhances overall user experience. Transparency ensures that users are informed not only about the robot's intentions, but also about its capabilities and the appropriate contexts for its use. For example, a household robot that communicates its cleaning schedule can help users to plan their activities around the robot's operations or prompt them to modify the robot's schedule. Transparency is not only beneficial for end-users but also advantageous for robot developers since it can help them design, test, and debug robots in real time \cite{theodorou2017designing}. For instance, a cleaning robot could explain the sequence of tasks it has or has not performed, so that the designer can obtain insights into its behavior, to be able to determine whether a specific action should or should not be part of a respective action sequence. Accordingly, the robot's functionalities can be readjusted, if this is required. Transparency in robots can also serve as a tool for developers and researchers to better understand and improve robot's functioning \cite{rahwan_machine_2019}. However, transparency can be viewed as a double-edged sword: While it can enhance trust and acceptance, it may also result in negative consequences, such as overtrust \cite{ososky2014determinants} or information overload, for instance, when explanations provided by the robot are too extensive \cite{ezenyilimba2023impact}. Therefore, while striving for transparency, it is imperative to balance the amount and type of information disclosed by the robot to prevent these potential drawbacks.

The necessity of transparency is also underscored in Article 13 of the EU AI Act, according to which ``AI systems shall be designed and developed in such a way to ensure that their operation is sufficiently transparent to enable providers and users to reasonably understand the system’s functioning'' \cite{prifti2024towards}. This emphasis on AI systems extends to the field of HRI, but despite the recognized significance of transparency in this domain, a clear definition is still missing. The absence of a consensual definition may hinder or compromise the applicability of regulations encouraging or enforcing transparency in robotic systems. Transparency as a construct has been widely discussed in prior research, yet this consensual definition of transparency remains to be formulated. Some authors define transparency in the sense of predictability, such as ``Transparency is essentially the opposite of unpredictability'' \cite[p. 193]{miller2014delegation}. Accordingly, if a robot provides cues that support its users in anticipating its next actions and states, the robot should be deemed more predictable and, thus, transparent. In contrast, other definitions of transparency emphasize the aspect of legibility. Wortham et al. \cite[p. 274]{wortham2017robot} defined transparency as ``{...} the extent to which the robot's ability, intent, and situational constraints are understood by users''. According to Kim and Hinds \cite[p. 81]{kim2006should}, the notion of explainability is also intertwined with transparency. They propose ``Transparency is the robot offering explanations of its actions''. These explanations can be provided before an action is performed, during an action, or after performing an action \cite{anjomshoae_explainable_2019, stange_explaining_2021}. Recent works, however, combine the aforementioned aspects, suggesting that transparency in HRI should encompass what a robot is doing, why it is doing that, and what it will do next \cite{angelopoulos2023robot,endsley2017here,alonso2018system}. Current theorizing on transparency in HRI suggests a multifaceted approach to transparency, integrating elements of explainability, legibility, and predictability. Such approach will foster a comprehensive understanding of a robotic system in HRI.

Above and beyond the lack of a consensual definition, \cite{schott2023literature} have highlighted the absence of consensus regarding the reliability and validity of metrics employed to assess the transparency of a robot's behavior. In addition, \cite{claure2022fairness} underscored the critical need for developing a standardized measurement framework to assess transparency, see also \cite{bartneck2009measurement}. When measuring transparency, straightforward questions such as ``Is the robot transparent?'' or ``Is the robot predictable?'' might seem adequate at first glance \cite{wengefeld2020laser,olatunji2021levels}. However, such face-valid, single-item measures fail to comprehensively capture all facets of a complex construct like transparency, leading to inconsistent and unreliable data \cite{schrum_concerning_2023}. The lack of consensual measurement for transparency in HRI research suggests that it may be understood differently from one individual to another. Similarly, laypersons may perceive transparency differently from HRI researchers and practitioners, who distinguish explainability, legibility, and predictability as the dimensions of transparency. For instance, legibility and predictability are not easily distinguishable in terms of human understanding. Indeed, human cognition deeply relies on predictive processes to understand the environment and prepare individuals for actions \cite{clark_whatever_2013, vesper_joint_2017}. It is then possible that people, when understanding something, have the feeling that it is predictable as well \cite{kahneman_noise_2021}. The development of a validated scale is, therefore, essential for methodological rigor and for the theoretical understanding of transparency.

\begin{figure}[h]
\centering
\includegraphics[width=\linewidth]{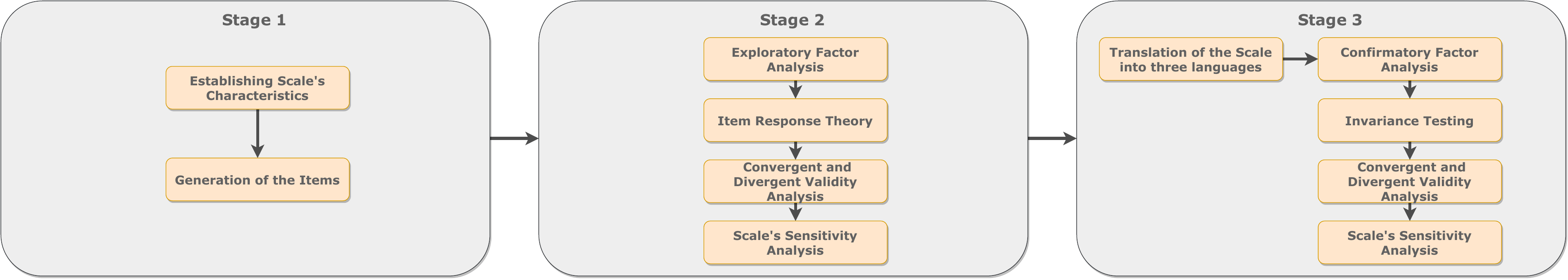}
\caption{The three stages of the scale development.}
\label{fig:methods}
\end{figure}

The development and use of a validated scale to measure the transparency of a robot's behavior represent a significant advancement in the field. It ensures valid and reliable measurements, facilitates cross-study comparisons, and helps roboticists achieve an optimal balance of transparency since, as previously discussed, it can have both positive and negative effects. 
To the best of our knowledge, this paper proposes the first measurement instrument to assess the perceived transparency of a robot's behavior. We present the validation process, which follows the state-of-the-art for scale creation \cite{boateng_best_2018, schrum_concerning_2023}. Specifically, the process comprises three distinct stages, as shown in Figure \ref{fig:methods}, each integral to ensure the robustness and validity of the resulting final scale. These stages consist of (1) Initial item generation to formulate a set of items, (2) Exploratory Factor Analysis (EFA) to discern a factor arrangement, and (3) a Confirmatory Factor Analysis (CFA) to validate the scale's factor structure. The research protocol for the work presented in this paper received approval from the Ethics Committee of Bielefeld University, under application number 2023-349, on December 12th, 2023.

%% file: 2_item_generation.tex
Stage 1 of the scale development involved an examination of existing definitions and measures to allow the creation of the first version of a scale assessing a robot's perceived transparency. A literature review suggested that a robot's behavior requires three characteristics to be transparent: Explainability, legibility, and predictability. However, transparency also emerges from how these dimensions help the human perceiver to build a mental model of the robot's functioning \cite{lee2023here,angelopoulos2023using}. We refer to this fourth aspect as meta-understanding. With these four characteristics established, we generated items that effectively capture each aspect of transparency. Here, we were guided by the methodological recommendation to generate three to five times the number of items intended for a resulting measurement instrument, with at least four items per dimension \cite{boateng_best_2018}. Consequently, 64 items in total were initially developed and distributed evenly across the four dimensions. Moreover, since no validated transparency scale is available in the literature yet, the first version of the scale was designed so items from previous studies that measure specific aspects of transparency (e.g., \textit{``I find the robot’s behavior easy to understand''}) were incorporated, as explained in recent HRI and Human-Computer Interaction (HCI) literature \cite{schott2023literature,ehsan2021expanding,alonso2018system}. Each item was formulated to carefully match the theoretical notions associated with explainability, legibility, predictability, and meta-understanding. This way, we made sure that the items were relevant and sufficient to fully capture each dimension. 

Furthermore, in Stage 1, we deliberately created multiple items with similar content or phrasing. The rationale behind this strategy is twofold: First, it allows for a more comprehensive coverage of the construct's content domain, ensuring that various nuances and aspects of each dimension are captured. Second, it provides the opportunity to empirically evaluate which items perform best in terms of psychometric properties during subsequent phases of scale refinement. By including redundant items at this stage (e.g., \textit{``The robot’s behavior is predictable''}, \textit{``The robot’s actions are unpredictable''}), we increase the likelihood of identifying the most robust and discriminating items for the final scale, potentially improving its overall reliability and validity \cite{devellis2021scale}.

In addition, to capture the different ways in which individuals might perceive and describe their experiences with robots, a mix of personal and non-personal statements was developed. Personal statements (e.g., \textit{``I can [...]'', ``I find [...]'', ``I feel [...]''}) refer to direct expressions of the participant's experiences and feelings. Non-personal statements (e.g., \textit{``The robot's behavior [...]'', ``The robot's explanation [...]'', ``The robot's actions [...]''}) provide an assessment of the robot's characteristics. Moreover, we integrated three reverse-coded items (asking for agreement with negatively phrased statements), each for personal and non-personal statements (e.g., \textit{``I find it difficult [...]'', ``The robot does not [...]''}) to counter-response biases and increase the reliability of responses \cite{sonderen2013ineffectiveness}. Responses are measured using a 7-point Likert scale, ranging from \textit{``Strongly Disagree'' (1)} to \textit{``Strongly Agree'' (7)}. This was done because 7-point Likert scales offer an optimal balance between ease of use, adjustment to memory span, and accuracy \cite{taherdoost2019best}. The initial 64 items can be found in Table S1 of the Supplementary Information.


%% file: 3_experiment1.tex
Following item generation, Stage 2 of the scale development and validation process featured an empirical experimental study. This enabled an exploratory factor analysis to examine the underlying structure of the scale. Such analysis served to confirm the hypothesized four-factor model of perceived transparency. EFA was used since it is instrumental in assessing the scale's factorial structure and facilitates item reduction by identifying items that strongly contribute to each factor. At this stage, an assessment of the scale items' difficulty and discrimination parameters was also included. Additionally, Stage 2 serves to evaluate the scale's convergent and divergent validity through controlled manipulations of the primary dimensions of transparency (explainability, legibility, and predictability), as explained in previous works \cite{angelopoulos2023robot,endsley2017here,alonso2018system}, with image vignettes (controlled visual scenarios). This process is crucial in order to confirm whether changes in these dimensions correspond to variations in perceived transparency. That way, we can provide evidence for the scale's sensitivity and construct validity.

Before Experiment 1, we conducted a pretest to identify a hypothetical everyday life scenario featuring HRI that would effectively discriminate transparency. This way, we maximize the scale's sensitivity and effectiveness for measuring transparency. Based on the pretest's findings, we selected a scenario where a robot heading towards a charging station to refill its battery (the results of the pretest can be found in the Supplementary Information).  

Experiment 1 employed a $2 \times 2 \times 2$ between-subject design which manipulated the explainability, legibility, and predictability of a robot's behavior as either low or high, using the pretested eight image vignettes. Experiment 1 was implemented on Qualtrics. Participants were presented with the purpose and procedure of the experiment. Those who gave their informed consent were randomly assigned to one of the eight between-subject conditions resulting from the manipulation of three variables related to transparency. After viewing the scenario, participants were required to self-report the following dependent variables: Perceived transparency with 64 items developed during Stage 1 (Item Generation), trust towards robots with 20 items from the Multi-Dimensional Measure of Trust (MDMT) \cite{ullman_measuring_2019}, and acceptance of robots using seven subscales from a toolkit based on the Unified Theory of Acceptance and Use of Technology (UTAUT), suggested for HRI research \cite{heerink_measuring_2009}. Additionally, demographic questions (i.e., age, gender, education, self-assessed English language proficiency) and prior experience with robots were assessed using a scale based on \cite{reich-stiebert_learning_2015}. Finally, two attention checks were included, one at the beginning and one at the end of the study. Only complete datasets from participants over 18 years of age and with a self-declared English proficiency at the A2 level (Elementary) and above were included. Data from people failing both attention checks were excluded. The data collection was planned to conclude after obtaining complete datasets from 320 participants. This sample size was strategically chosen to meet the requirement of having at least 5 participants per item for factor analysis \cite{hatcher1994step}, considering the transparency scale consisted of 64 items. Additionally, the sample size met the recommended threshold of 300 participants for robust factor analysis \cite{tabachnick_using_2019}. Participants for Experiment 1 were recruited via Prolific, and they were reimbursed with £1.50 for participating. The pre-registration for Experiment 1 is available at \url{https://aspredicted.org/ZTT_STV}.

\subsection*{Sample}
For Experiment 1, we recruited 371 participants. Following our pre-registered exclusion criteria, 49 participants were removed from the sample; 13 failed to complete the study, and 36 were excluded for not passing both attention checks. Thus, the final sample comprised $N=322$ participants, slightly exceeding our pre-registered target by two participants. The demographic breakdown, as detailed in Table S3 of the Supplementary Information, was as follows: 182 females, 135 males, two identifying as diverse, two who preferred not to disclose their gender, and one gender-fluid participant. The age range of participants was between 18 and 71 years old ($M=29.820$, $SD=9.340$).

\subsection*{Results}

An Exploratory Factor Analysis using \textit{SPSS 28.0} was employed to assess the factorial validity of the 64-item scale. This analysis utilized Principal Axis Factoring (PAF) with an oblique (direct-oblimin) rotation, which allows intercorrelations among factors \cite{sass_comparative_2010, fabrigar_evaluating_1999}. This approach was accompanied by evaluations of univariate statistics, the Kaiser-Meyer Olkin (KMO) measure of sampling adequacy, Bartlett's test of sphericity, eigenvalues, and the Scree plot.

\begin{figure}[h]
\centering
\includegraphics[width=0.5\linewidth]{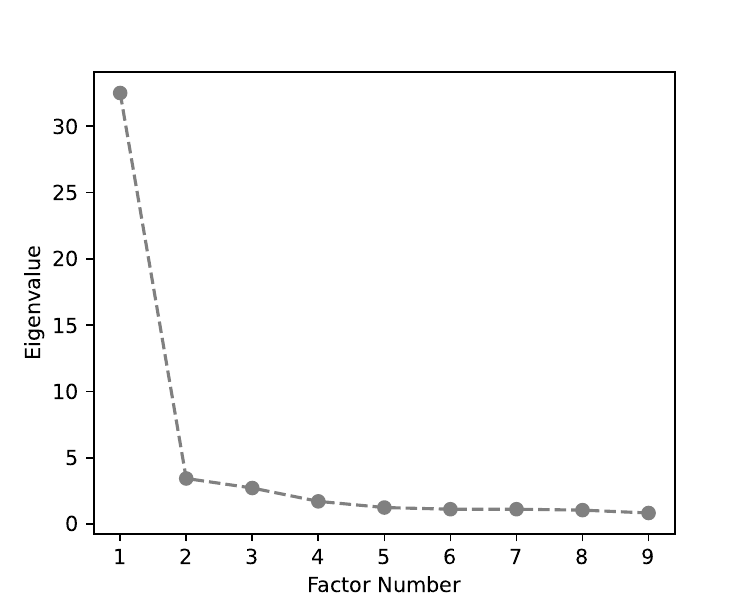}
\caption{Scree plot illustrating the eigenvalues of extracted factors.}
\label{fig:scree}
\end{figure}

The KMO for the sample data was $.976$. Accordingly, the data appeared suitable for an EFA \cite{kaiser_index_1974, shrestha_factor_2021}. Bartlett's test further validated the appropriateness of this statistical method, yielding a $\chi^2 = 18084.169$, $\text{df} =2016$, $p<.001$. The Scree plot, in Figure \ref{fig:scree}, revealed a clear breakpoint after five factors, all of which accounted for $64.154\%$ of the variance. This result demonstrated the strong explanatory power of the scale.

Initially, all 64 items were retained for the initial EFA. The results, detailing the factor loadings, are depicted in Tables S5 and S6 of the Supplementary Information. Subsequent iterations of the EFA were conducted using the same settings, with item reduction guided by specific criteria to ensure a robust factor structure. These criteria involved removing items with loadings below $.400$ to maintain a strong factor representation and enhance interpretability, following the recommendations by \cite{guvendir2022item}. Additionally, items loaded on multiple factors with a loading difference of less than $.100$ were excluded to minimize ambiguity in item association, following \cite{guvendir2022item}. Furthermore, factors containing fewer than three items were removed to ensure reliable and meaningful factor solutions, adhering to the guidelines proposed by \cite{knekta2019one, schrum_concerning_2023}.

Initial item removals included five rounds of EFA, in which low-factor loadings, multiple high-factor loadings, and factor loadings on underrepresented factors were taken into account as exclusion criteria. The final EFA resulted in a refined 31-item scale structured into three distinct factors. The revised factor model demonstrates robust factor loadings. The detailed factor loadings for this definitive structure are presented in Tables S7 and S8 of the Supplementary Information.

 Following the iterations of EFA, an inter-item correlation analysis was conducted to minimize redundancy within each factor. Items exhibiting excessively high correlations (greater than $.700$ with more than one item) were removed to ensure the distinctiveness of each item within the factors. This approach confirmed that the remaining items clearly and distinctly represented the underlying constructs without undue overlap. At first, the inter-item correlation at Factor 1 was analyzed, having correlation coefficients ranging from $.511$ to $.793$ with a mean of $M=.680$. Table S9 of the Supplementary Information shows that four items had to be removed. Table S10 of the Supplementary Information shows the final inter-item correlation matrix of Factor 1 ranging from $.511$ to $.728$ with a mean of $M=.642$. Afterward, we analyzed Factor 2 (as shown in Table S11 of the Supplementary Information), in which the correlation coefficients ranged from $.479$ to $.740$ with $M=.607$. One item had correlation coefficients greater than .$700$ with more than one item and was removed. A new analysis was conducted (depicted in Table S12 of the Supplementary Information), showing that the correlation coefficients ranged from $.479$ to $.690$ with a mean of $M=.598$. Finally, as depicted in Table S13 of the Supplementary Information, the values of the inter-item correlation for Factor 3 ranged from $.490$ to $.683$ with a mean of $M=.572$. No items were removed, leading to a 26-item scale.
 
To validate the robustness and appropriateness of the 26-item scale's factor structure, we conducted a series of statistical analyses, including tests of sampling adequacy, factorability, internal consistency, and convergent validity. The KMO reaffirmed the suitability of the data for EFA with a value of $.968$. Bartlett's test supported the factorability of the correlation matrix, yielding a $\chi^2 = 6217$, $\text{df} =325$, $p<.001$. The new factor loadings for both the pattern and structure matrices were examined to confirm the appropriateness of the items within each factor. Internal consistency was assessed using Cronbach's Alpha ($\alpha$) and McDonald's Omega ($\omega$). All factors demonstrated high internal consistency, with Factor 1 recording values of $.930$ for both Cronbach's Alpha and McDonald's Omega, Factor 2 showing a Cronbach's Alpha of $.926$ and McDonald's Omega of $.927$, and Factor 3 similar to Factor 1 with values of $.930$ for both metrics. Composite Reliability (CR) and Average Variance Extracted (AVE) were calculated to evaluate the factors' convergent validity. The CR values for Factor 1, Factor 2, and Factor 3 were satisfactory at $.800$, $.800$, $.735$, respectively. Regarding the AVE, Factor 1, Factor 2, and Factor 3 recorded values of $.438$, $.438$, and $.463$, respectively. Despite the AVE values being marginally below the ideal threshold of $.5$, the high CR values justify proceeding with the utilization of the scale \cite{rahman2018structural}. The comprehensive results from these evaluations are presented in Table \ref{tab:finalsetting26efa}.


\begin{table}[h]

\begin{threeparttable}
\setlength{\tabcolsep}{4pt} 
\begin{tabular}{@{} l S[table-format=-1.3] l S[table-format=-1.3] c c c c c @{}} 

\toprule

\multicolumn{1}{c}{\small{Item}} & \multicolumn{3}{c}{\small{Factor Pattern Loadings}} & {\small{\makecell{Factor Structure \\ Loadings}}} & \multicolumn{2}{c}{\small{\makecell{Internal \\ Reliability}}} & \multicolumn{2}{c@{}}{\small{\makecell{Convergent \\ Validity}}} \\ 
\cmidrule(lr){2-4} \cmidrule(lr){6-7} \cmidrule(lr){8-9}
& \footnotesize{Factor 1} & \footnotesize{Factor 2} & \footnotesize{Factor 3} & & \footnotesize{$\alpha$} & \footnotesize{$\omega$} & \footnotesize{CR} & \footnotesize{AVE} \\
\midrule

\footnotesize{The robot's overall functioning is a mystery to me.}                          & \textbf{-.797} & {-.008} & {.032}  & {-.683}                     & \multirow{10}{*}{.930}  & \multirow{10}{*}{.930} & \multirow{10}{*}{.800}  & \multirow{10}{*}{.438} \\

\footnotesize{It is hard to make sense of the robot’s general functioning.}                 & \textbf{-.770} & {-.032} & {.058}  & {-.656}      \\

\footnotesize{It is difficult to get a clear picture of the robot's overall operations.}    & \textbf{-.706} & {-.101} & {.021}  & {-.690}       \\

\footnotesize{I am confused about the robot's general objectives.}                          & \textbf{-.692} & {-.144} & {.020}  & {-.715}            \\

\footnotesize{I am unsure what the robot does.}                                             & \textbf{-.688} & {-.098} & {-.038}  & {-.724}            \\

\footnotesize{I cannot comprehend the robot’s inner processes.}                             & \textbf{-.673} & {.004} & {-.056}  & {-.641}            \\

\footnotesize{I cannot explain the robot’s behavior.}                                       & \textbf{-.671} & {.135} & {-.284}  & {-.729}            \\

\footnotesize{It is impossible to know what the robot does.}                                & \textbf{-.638} & {-.049} & {-.124}  & {-.715}            \\

\footnotesize{It is clear to me what the robot does.}                                       & \textbf{.467} & {.347} & {.106}  & {.799}            \\

\footnotesize{I have a clear understanding of how the robot operates in general.}           & \textbf{.426} & {.320} & {.067}  & {.739}            \\

\midrule

\footnotesize{I feel like the robot's explanations are useful.}                             & {.024} & \textbf{.832} & {-.039}  & {.688}                      & \multirow{7}{*}{.926}  & \multirow{7}{*}{.927} & \multirow{7}{*}{.865}   & \multirow{7}{*}{.481} \\

\footnotesize{The robot explains complex tasks in a way that is easy to understand.}        & {.119} & \textbf{.753} & {-.024}  & {.718}        \\

\footnotesize{The robot provides detailed explanations of its actions.}                     & {.008} & \textbf{.708} & {.084}  & {.678}          \\

\footnotesize{The robot provides clear explanations for its actions.}                       & {.067} & \textbf{.657} & {.163}  & {.757}          \\

\footnotesize{The robot's explanations for its actions are straightforward.}                & {.049} & \textbf{.654} & {.223}  & {.792}          \\

\footnotesize{I feel informed about the robot's activities.}                                & {.251} & \textbf{.625} & {.044}  & {.787}          \\

\footnotesize{The robot conveys its overall state effectively.}                             & {.096} & \textbf{.600} & {.110}  & {.688}          \\

\midrule

\footnotesize{It is easy for me to foresee the robot's future actions.}                     & {-.043} & {-.011} & \textbf{.829}  & {.685}          
 & \multirow{9}{*}{.930}  & \multirow{9}{*}{.930} & \multirow{9}{*}{.735} & \multirow{9}{*}{.463} \\

\footnotesize{The robot's behavior is predictable.}                                         & {-.034} & {.090} & \textbf{.785}  & {.739}          \\

\footnotesize{I feel confident in predicting the robot's next moves.}                       & {.057} & {.064} & \textbf{.732}  & {.751}              \\        

\footnotesize{It is easy to anticipate what will follow the robot’s behavior.}              & {.057} & {.038} & \textbf{.723}  & {.721}             \\

\footnotesize{It is difficult for me to tell what the robot will do next.}                  & {-.318} & {.168} & \textbf{-.666}  & {-.728}             \\

\footnotesize{The robot's next steps are clear to me.}                                      & {.049} & {.209} & \textbf{.656}  & {.798}             \\

\footnotesize{The robot's actions are obvious.}                                             & {-.018} & {.266} & \textbf{.585}  & {.725}             \\

\footnotesize{The robot provides cues that help predict its next actions.}                  & {-.082} & {.336} & \textbf{.554}  & {.700}             \\

\footnotesize{The robot's behavior does not help predict what it will do next.}             & {-.277} & {.076} & \textbf{-.536}  & {-.654}             \\

\bottomrule
\end{tabular}

\caption{The scale after the removal of items with low loadings and high inter-item correlation.}
\label{tab:finalsetting26efa}
\end{threeparttable}
\end{table}

\begin{figure}[h]
\centering
\includegraphics[width=0.5\linewidth]{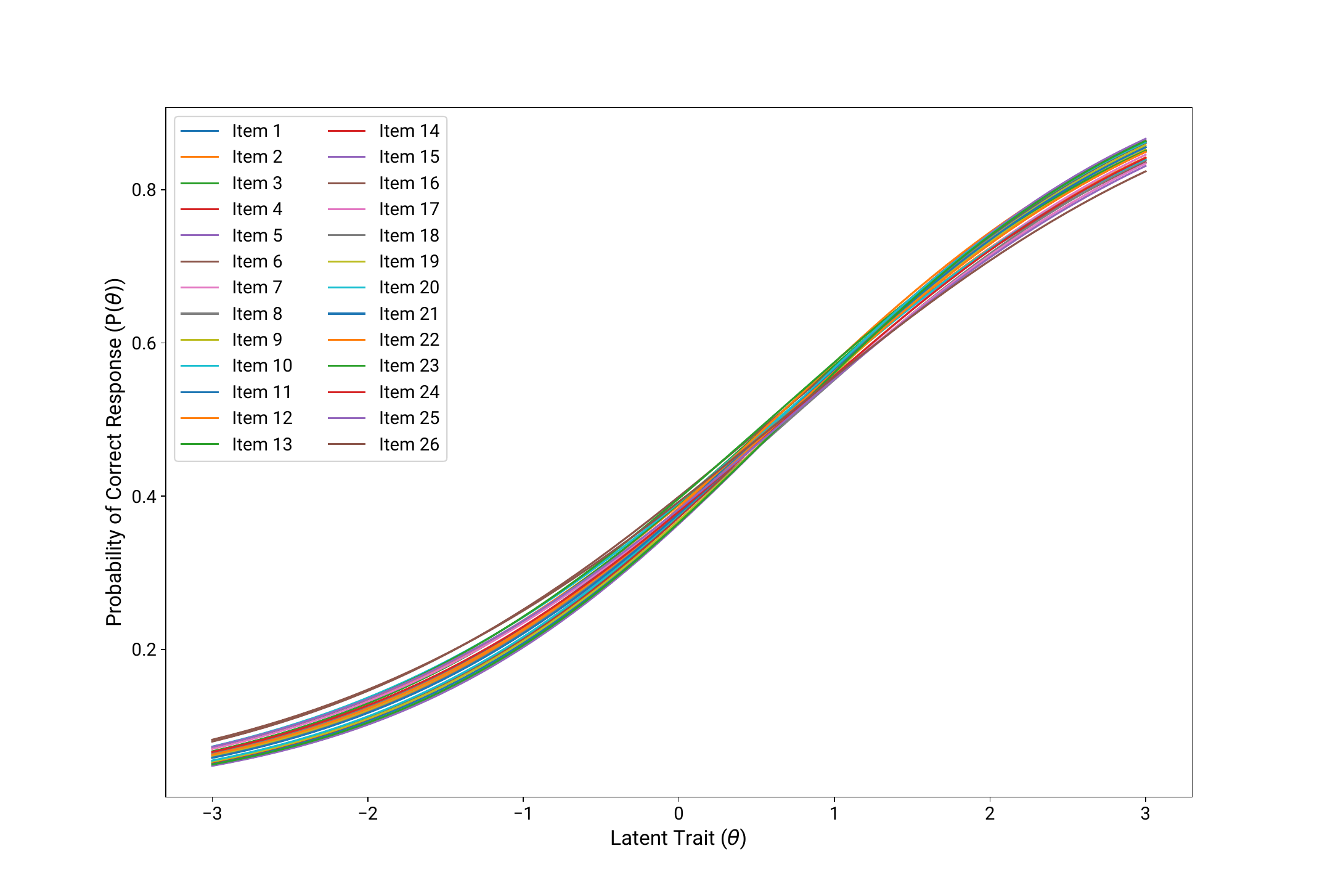}
\caption{The Item Characteristic Curve exhibits a sigmoid shape.}
\label{fig:icc}
\end{figure}

Moreover, to further evaluate the 26-item scale, we employed Item Response Theory (IRT) to analyze the difficulty and discrimination of each item \cite{lalor2016building}, as depicted in Table S14 of the Supplementary Information. IRT is a statistical approach that models the relationship between an individual's response to an item and their level of the underlying construct being measured, known as the latent trait. The IRT model results are presented in Figure \ref{fig:icc} using an Item Characteristic Curve (ICC) for all 26 items. Such ICC featured a sigmoid shape, indicating a good fit between the model and the empirical data \cite{lalor2016building}. Specifically, the sigmoid pattern suggests that as the latent trait increases, the probability of a correct response also increases systematically, reflecting an effective measurement scale. For item difficulty, values ranged from $.580$ to $.710$ ($M=.660$, $SD=.040$). Regarding item discrimination, the range was between $.660$ and $0.810$ ($M=.745$, $SD=.045$). From this follows that the items have moderate difficulty levels and moderate to high discrimination power, suggesting they are effective in differentiating between individuals.

 Prior to assessing the sensitivity of the 26-item scale resulting from our factor analysis for the actual transparency of robot behaviors, we assessed the reliability of the questionnaires (i.e., MDMT, UTAUT, and experience with robots) through Cronbach's alpha, confirming their internal consistency ($>.7$) as demonstrated in Table S15 of the Supplementary Information. Afterward, we conducted a series of 2x2x2 factorial ANCOVAs. These analyses incorporated manipulated levels of explainability, legibility, and predictability (each set at low versus high) as the independent variables while utilizing the means for each factor as well as the mean of all items from the 26-item scale as the dependent variables.

\begin{figure}[h]
\centering
\includegraphics[width=0.7\linewidth]{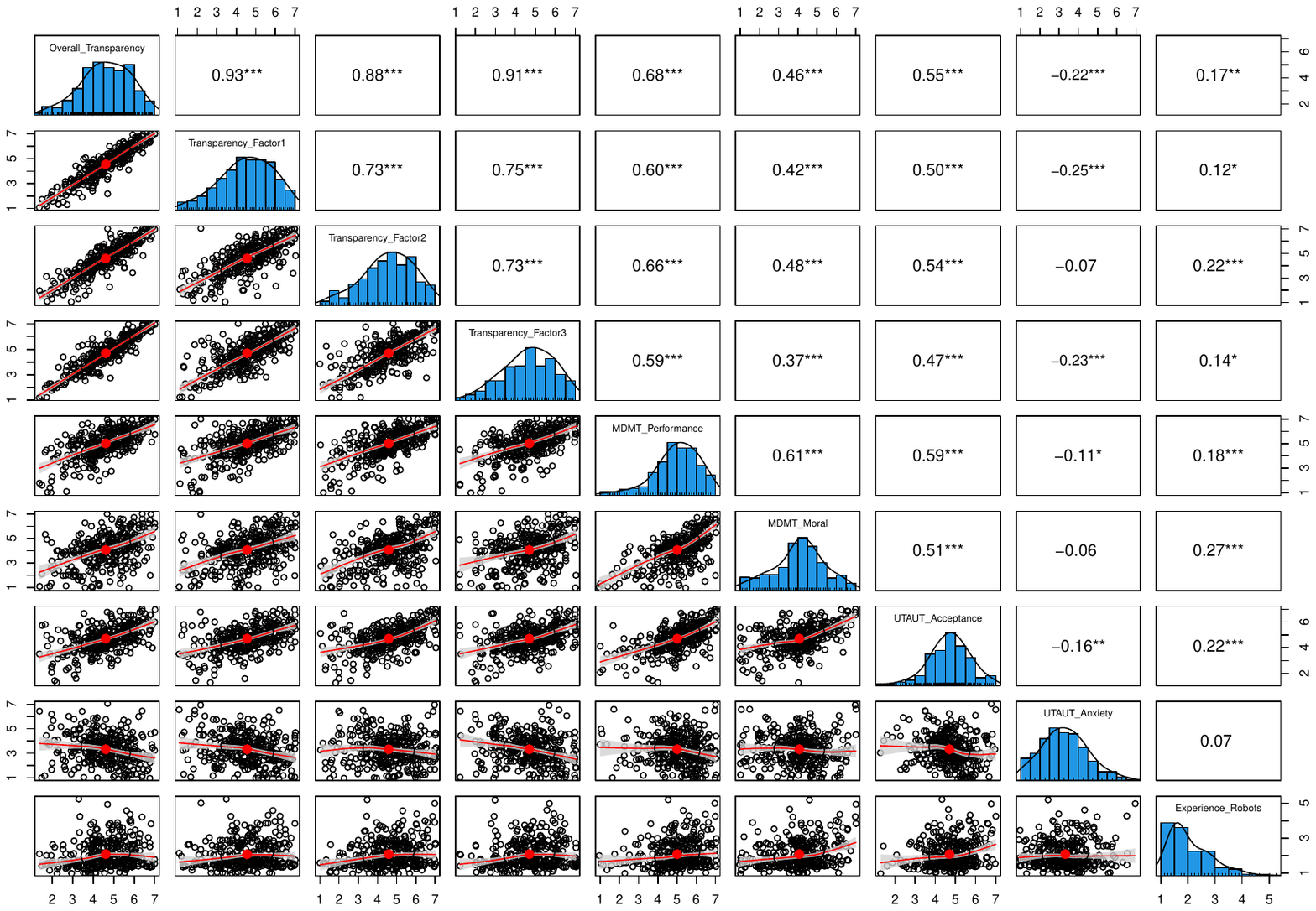}
\caption{Correlation matrices between the dependent and control variables of Experiment 1.}
\label{fig:correlation}
\end{figure}

In cases where experience with robots was significantly correlated with the tested dependent variable (see Figure \ref{fig:correlation}), it was included as a covariate in the ANCOVA. Table \ref{tab:CombinedANOVA1} shows the results of all the ANCOVAs. We found a significant main effect of manipulated explainability, legibility, and predictability on transparency calculated using the average of all the items. However, no interaction effect between manipulated explainability, legibility, and predictability was identified. Similarly, we obtained significant main effects of manipulated explainability, legibility, and predictability on Factor 1, Factor 2, and Factor 3. However, no interaction effect between manipulated explainability, legibility, and predictability on Factor 1, Factor 2, or Factor 3 was observed. Prior experience with robots was a significant covariate for average perceived transparency, Factor 1, Factor 2, and Factor 3.

Regarding trust towards the robot, we found significant main effects of manipulated explainability and legibility on performance (see Table S16 of the Supplementary Information) and moral trust towards the robot (see Table S17 of the Supplementary Information). However, although the main effect of manipulated predictability on performance trust was significant, we did not find a significant effect of manipulated predictability on moral trust towards the robot. No significant interaction effect between manipulated explainability, legibility, and predictability on performance trust or moral trust was discovered. Prior experience with robots was a significant covariate for performance trust and moral trust towards the robot. Finally, we observed a significant main effect of manipulated explainability, legibility, and predictability on acceptance of the robot (see Table S18 of the Supplementary Information). Prior experience with robots was a significant covariate for acceptance. No interaction effect between manipulated explainability, legibility, and predictability on acceptance was identified. No significant effect of the independent variables on anxiety was found (see Table S19 of the Supplementary Information).

\begin{table}[h!]
\centering
\begin{tabular}{lccccccc}
\hline
\multicolumn{1}{c}{Parameter}            & df  & SS     & MS     & $F$ value & $p$ value            & $\eta p^2$   & 95\% CI         \\ \hline
\multicolumn{8}{c}{\textbf{Dependent variable: Overall Score of the Transparency}} \\ \hline
Explainability condition                 & 1   & 74.10 & 74.13 & 70.53   & \textless .001*** & .18            & {[}0.12, 1.00{]} \\
Legibility condition                     & 1   & 9.20  & 9.23  & 8.78    & .003**             & .03            & {[}0.01, 1.00{]} \\
Predictability condition                 & 1   & 27.70 & 27.67 & 26.33   & \textless .001***    & .08            & {[}0.04, 1.00{]} \\
Prior experience with robots             & 1   & 17.60 & 17.64 & 16.78   & \textless .001***    & .05            & {[}0.02, 1.00{]} \\
Explainability: Legibility                & 1   & 0.70  & 0.70  & 0.67    & .415               & \textless .01 & {[}0.00, 1.00{]} \\
Explainability: Predictability            & 1   & 0.30  & 0.29  & 0.28    & .599               & \textless .01 & {[}0.00, 1.00{]} \\
Legibility: Predictability                & 1   & 1.90  & 1.90  & 1.803   & .180               & .01            & {[}0.00, 1.00{]} \\
Explainability: Legibility: Predictability & 2   & 0.60  & 0.62  & 0.59    & .444               & \textless .01  & {[}0.00, 1.00{]} \\
Residuals                                & 313 & 329.0 & 1.05  &         &                    &                &                  \\ \hline
\multicolumn{8}{c}{\textbf{Dependent variable: Factor 1}} \\ \hline
Explainability condition                 & 1   & 67.50  & 67.50 & 46.55   & \textless .001*** & .13            & {[}0.08, 1.00{]} \\
Legibility condition                     & 1   & 13.30  & 13.35 & 9.21    & .003**             & .03            & {[}0.01, 1.00{]} \\
Predictability condition                 & 1   & 31.80  & 31.85 & 21.96   & \textless .001*** & .07            & {[}0.03, 1.00{]} \\
Prior experience with robots             & 1   & 12.50  & 12.48 & 8.60    & .004**             & .03            & {[}0.01, 1.00{]} \\
Explainability: Legibility                & 1   & 0.90   & 0.94  & 0.65    & .421               & \textless .01 & {[}0.00, 1.00{]} \\
Explainability: Predictability            & 1   & 0.10   & 0.07  & 0.05    & .823               & \textless .01 & {[}0.00, 1.00{]} \\
Legibility: Predictability                & 1   & 2.80   & 2.83  & 1.95    & .163               & .01            & {[}0.00, 1.00{]} \\
Explainability: Legibility: Predictability & 2   & 1.60   & 1.58  & 1.09    & .298               & \textless .01  & {[}0.00, 1.00{]} \\
Residuals                                & 313 & 453.90 & 1.45  &         &                    &                &                  \\ \hline
\multicolumn{8}{c}{\textbf{Dependent variable: Factor 2}} \\ \hline
Explainability condition                 & 1   & 116.50 & 116.45 & 91.57   & \textless .001*** & .23            & {[}0.16, 1.00{]} \\
Legibility condition                     & 1   & 6.60   & 6.57   & 5.17    & .024*              & .02            & {[}0.00, 1.00{]} \\
Predictability condition                 & 1   & 17.30  & 17.32  & 13.62   & \textless .001*** & .04            & {[}0.01, 1.00{]} \\
Prior experience with robots             & 1   & 32.60  & 32.58  & 25.61   & \textless .001*** & .08            & {[}0.04, 1.00{]} \\
Explainability: Legibility                & 1   & 1.70   & 1.73   & 1.36    & .244               & \textless .01 & {[}0.00, 1.00{]} \\
Explainability: Predictability            & 1   & 0.10   & 0.12   & 0.10    & .756               & \textless .01 & {[}0.00, 1.00{]} \\
Legibility: Predictability                & 1   & 1.00   & 1.00   & 0.78    & .377               & \textless .01 & {[}0.00, 1.00{]} \\
Explainability: Legibility: Predictability & 2   & 0.10   & 0.06   & 0.05    & .824               & \textless .01  & {[}0.00, 1.00{]} \\
Residuals                                & 313 & 398.10 & 1.27   &         &                    &                &                  \\ \hline
\multicolumn{8}{c}{\textbf{Dependent variable: Factor 3}} \\ \hline
Explainability condition                 & 1   & 54.00  & 54.04 & 41.52   & \textless .001*** & .12            & {[}0.07, 1.00{]} \\
Legibility condition                     & 1   & 7.40   & 7.42  & 5.70    & .018*              & .02            & {[}0.00, 1.00{]} \\
Predictability condition                 & 1   & 32.40  & 32.37 & 24.87   & \textless .001*** & .07            & {[}0.03, 1.00{]} \\
Prior experience with robots             & 1   & 14.20  & 14.20 & 10.91   & .001**             & .03            & {[}0.01, 1.00{]} \\
Explainability: Legibility                & 1   & 0.10   & 0.10  & 0.08    & .781               & \textless .01 & {[}0.00, 1.00{]} \\
Explainability: Predictability            & 1   & 2.50   & 2.52  & 1.94    & .165               & .01            & {[}0.00, 1.00{]} \\
Legibility: Predictability                & 1   & 1.80   & 1.77  & 1.36    & .244               & \textless .01 & {[}0.00, 1.00{]} \\
Explainability: Legibility: Predictability & 2   & 0.50   & 0.46  & 0.36    & .551               & \textless .01  & {[}0.00, 1.00{]} \\
Residuals                                & 313 & 407.40 & 1.30  &         &                    &                &                  \\ \hline
\end{tabular}
\begin{tablenotes}
\item Note: df = Degrees of freedom; SS = Sum of Squares; MS = Mean Squares;\\ *p< .05; **p< .01; ***p< 0.001;
\end{tablenotes}
\caption{Results of the 3-way ANOVA with the manipulation of explainability, legibility, and predictability of the robot's behavior as independent variables}
\label{tab:CombinedANOVA1}
\end{table}

%% file: 4_experiment2.tex
Following Stage 2, which concluded with a 26-item scale, we proceeded to Stage 3 of the scale development process. In this stage, we designed Experiment 2 to validate the factor structure. The first step involved translating the 26-item scale into German and Italian, employing a forward and backward translation process by two independent translators, as explained in \cite{fenn2020development}.  This was essential for testing the scale's properties across languages and for ensuring cross-linguistic reliability and validity. Additionally, the effectiveness of the scale was assessed by measuring participant responses under conditions of low versus high transparency with video vignettes. These vignettes were based on the scenarios of Stage 2, and the video format was meant to have the transparency of the robot assessed in more ecological settings (i.e., after observing a real robot in action). Afterward, Stage 2 included a CFA to validate the factor structure of the 26-item scale identified in the previous Stage. As a result, Stage 3 was critical in demonstrating the scale's capacity to discriminate between different levels of perceived transparency, demonstrating its practical applicability and psychometric consistency across diverse cultural contexts, akin to the validation shown in previous studies \cite{vagnetti2024instruments}. 

Experiment 2 was designed as a $2$ (low and high transparency conditions) $\times 3$ (3 languages) between-subject experiment where that served to manipulate the transparency of a robot's behavior via video vignettes featuring the robot Pepper (Softbank Robotics). Given that explainability, legibility, and predictability all had a significant influence on each factor of the scale in the previous stage, for Stage 2 we used two conditions to manipulate the overall transparency. Moreover, Experiment 2 was conducted in three languages (English, German, and Italian). Participants were presented with the purpose and procedure of the experiment. Those who gave informed consent were randomly assigned to one of the two transparency conditions, and the corresponding video was displayed. To ensure that participants watched the entire video featuring the robot behaving high vs. low in transparency at least once, the "Continue" button appeared only after approximately 5 seconds. Following the video, participants completed the resulting refined scale from Stage 2, followed by the MDMT scale \cite{ullman_measuring_2019} and the subscales from the UTAUT toolkit \cite{heerink_measuring_2009}. Translated versions of these scales, prepared by language experts, were used for German and Italian participants. Additionally, demographic questions (i.e., age, gender, education, self-assessed English language proficiency) and prior experience with robots were assessed using a scale based on\cite{reich-stiebert_learning_2015}. Finally, two attention checks were included at the study's beginning and end, and one memory check was included after the video's projection. Only complete datasets from participants over 18 years of age and with a self-declared language level (English, German, or Italian, depending on the language condition) above B2 level (Upper Intermediate) were included. Data from people failing the attention checks or the memory check were excluded. Following the recommendations from \cite{tabachnick_using_2019}, the data collection was planned to conclude after obtaining complete datasets from 300 participants per sub-sample, resulting in a total number of 900 complete data sets. The Study 2 pre-registration details are available at \url{https://aspredicted.org/JMC_3B8}.

\subsection*{Sample}

We recruited 927 participants using Prolific. Following our pre-registered exclusion criteria, 26 participants were disqualified; 17 failed to complete the study, 1 was excluded for not passing the video attention check, and 8
were removed for insufficient language proficiency. 
The final sample comprised $N=901$ participants. As detailed in Table S20 of the Supplementary Information, the demographics were as follows: 427 females, 452 males, 14 identifying as diverse, and 8 who preferred not to disclose their gender. The age range of participants was between 18 and 71 years ($M=37.607$, $SD=12.215$), and their experience with robots was rated below average ($M=1.930$, $SD=1.160$). The average duration of participation was recorded at 9.1 minutes, and each participant received compensation of £1.50 for their involvement in the study.

\subsection*{Results}

A Confirmatory Factor Analysis using the \textit{lavaan} package in R was employed to verify the factor structure of the 26-item scale. The CFA results for the English, German, and Italian versions of the scale indicated a good model fit for the English and Italian versions, with Comparative Fit Index (CFI) values of 0.930 and 0.943, respectively, and Tucker-Lewis index (TLI) values of 0.924 and 0.937. The Root Mean Square Error of Approximation (RMSEA) values were 0.073 for English and 0.067 for Italian, both within acceptable ranges. The Standardized Root Mean Square Residual (SRMR) values were 0.053 and 0.045, respectively, indicating a good fit. The German version showed weaker fit indices with a CFI of 0.890, TLI of 0.879, RMSEA of 0.093, and SRMR of 0.062, suggesting a less optimal fit than the other versions. Despite this, all factor loadings were significant across all three language versions, indicating that the items loaded well onto their respective factors. Table \ref{tab:standards} shows the specific analysis results.

\begin{table}[h]

\centering
\begin{threeparttable}
\begin{tabular}{@{} l c c c c c @{}} 
\toprule
{Standards} & {English} & {German} & {Italian} & {Acceptable} & {Excellent}\\ 

\midrule
Minimum fit function chi-square ($\chi^2$) & 774.253 & 1067.055 & 694.216 & - & -\\
Degrees of freedom ($df$) & 296 & 296 & 296 &  & \\
$\chi^2/df$ & 2.620 & 3.600 & 2.350 & $<5.00$ & $<3.00$\\
GFI & 0.828 & 0.756 & 0.841 & $>0.80$& $>0.90$\\
RMSEA & 0.073 & 0.093 & 0.067 & $<0.08$ & $<0.06$ \\
AGFI & 0.797 & 0.711 & 0.811 & $>0.80$ & $>0.90$\\
NFI & 0.892 & 0.854 & 0.905 & $>0.85$ & $>0.90$\\
CFI & 0.930 & 0.890 & 0.943 & $>0.90$ & $>0.95$\\
TLI & 0.924 & 0.879 & 0.937 & $>0.90$ & $>0.95$\\
IFI & 0.930 & 0.900 & 0.940 & $>0.90$ & $>0.95$\\
SRMR & 0.053 & 0.062 & 0.045 & $<0.08$ & $<0.05$\\
\bottomrule
\end{tabular}
\caption{Comparative Fit Indices for English, German, and Italian Models.}
\label{tab:standards}
\end{threeparttable}
\end{table}

Following the individual CFAs, measurement invariance testing was conducted to assess the comparability of the scale across the three language versions. Results indicated that the scale achieves configural and metric invariance, but not scalar or residual invariance. This suggests that while the construct and factor loadings are comparable across English, German, and Italian, there are differences in item intercepts and residual variances. Specifically, the configural invariance model, which tests if the basic structure is similar across groups, showed a good fit with a CFI of 0.910, TLI of 0.902, and RMSEA of 0.078, indicating that the factor structure is consistent across languages. The metric invariance model, which constrains factor loadings to be equal across groups, also showed acceptable fit (CFI of 0.906, TLI of 0.907, and RMSEA of 0.082), though the chi-squared difference test was significant, $\Delta \chi^2 = 120.47$, $p < 0.001$). 
The scalar invariance model, which additionally constrains item intercepts to be equal across groups, showed a significant deterioration in fit compared to the metric model, $\Delta \chi^2 = 283.93$, $p < 0.001$, although the overall fit indices remained similar (CFI = 0.906, TLI = 0.907, RMSEA = 0.082). This suggests that while the overall structure and factor loadings are comparable, there may be differences in how participants from different language groups interpret or respond to specific items on the scale. The residual invariance model, which further constrains residual variances to be equal across groups, likewise showed a significant decrease in fit, $\Delta \chi^2 = 226.99$, $p < 0.001$, with a slight drop in CFI to 0.898. This indicates that the unexplained variance in item responses could differ across language groups. Table \ref{tab:invariance} shows the specific analysis results.

\begin{table}[h]
\centering
\begin{threeparttable}
\begin{tabular}{lccccccccc}
\toprule
Model & $\chi^2$ & df & $\Delta \chi^2$ & $\Delta$df & CFI & TLI & RMSEA & SRMR \\
\midrule
Configural & 2535.5 & 888 &  &  & 0.910 & 0.902 & 0.078 & 0.071 \\
Metric & 2656.0 & 934 & 120.47* & 46 & 0.906 & 0.907 & 0.082 & 0.071 \\
Scalar & 2939.9 & 980 & 283.93* & 46 & 0.906 & 0.907 & 0.082 & 0.071 \\
Residual & 3166.9 & 1032 & 226.99* & 52 & 0.898 & 0.903 & 0.083 & 0.070 \\
\bottomrule
\end{tabular}
\begin{tablenotes}
\item Note: df = Degrees of freedom; $\Delta \chi^2$ = Change in Chi-Square; \\ $\Delta$df = Change in Degrees of Freedom\\ *p< .05; **p< .01; ***p< 0.001;
\end{tablenotes}
\caption{Invariance Testing Results for Configural, Metric, Scalar, and Residual Models}
\label{tab:invariance}
\end{threeparttable}
\end{table}

After examining the invariance, it was crucial to assess the scale's convergent validity and internal consistency across the three languages to ensure that the constructs are consistently and accurately measured. Internal consistency was robust across all languages, with Cronbach's Alpha values ranging from .920 to .951 and McDonald's Omega values from .940 to .960, showing high internal consistency for all factors. CR values for all factors exceeded the acceptable threshold of .70, ranging from .760 to .951, ensuring good convergent validity. Additionally, AVE values, which ranged from .547 to .736, consistently exceeded the .50 threshold, showing an improvement from Experiment 1 and confirming that a substantial portion of variance is explained by the factors across all languages. The results are depicted in Table \ref{tab:reliability_validity}.

\begin{table}[h]

\centering
\begin{threeparttable}
\begin{tabular}{@{} l c c c c c c c c c c c c c @{}} 

\toprule

\multicolumn{1}{c}{\small{Factor}} & \small{\makecell{No. of \\ Items}} & \multicolumn{6}{c}{\small{Internal Consistency}} & \multicolumn{6}{c}{\small{Convergent Validity}} \\ 
\cmidrule(lr){3-8} \cmidrule(lr){9-14}
& & \multicolumn{2}{c}{\footnotesize{English}} & \multicolumn{2}{c}{\footnotesize{German}} & \multicolumn{2}{c}{\footnotesize{Italian}} & \multicolumn{2}{c}{\footnotesize{English}} & \multicolumn{2}{c}{\footnotesize{German}} & \multicolumn{2}{c@{}}{\footnotesize{Italian}} \\ 
\cmidrule(lr){3-4} \cmidrule(lr){5-6} \cmidrule(lr){7-8} \cmidrule(lr){9-10} \cmidrule(lr){11-12} \cmidrule(lr){13-14}
& & \footnotesize{$\alpha$} & \footnotesize{$\omega$} & \footnotesize{$\alpha$} & \footnotesize{$\omega$} & \footnotesize{$\alpha$} & \footnotesize{} & \footnotesize{CR} & \footnotesize{AVE} & \footnotesize{CR} & \footnotesize{AVE} & \footnotesize{CR} & \footnotesize{AVE} \\ 
\midrule

\footnotesize{Factor 1} & 10 & .940 & .950 & .920 & .950 & .930 & .940 & .849 & .599 & .760 & .547 & .817 & .554 \\

\midrule

\footnotesize{Factor 2} & 7 & .945 & .946 & .945 & .946 & .951 & .953 & .945 & .712 & .875 & .590 & .951 & .736 \\

\midrule

\footnotesize{Factor 3} & 9 & .950 & .960 & .950 & .960 & .950 & .960 & .849 & .669 & .860 & .668 & .878 & .686 \\

\bottomrule
\end{tabular}

\caption{Internal consistency and convergent validity for each factor across the scale's English, German, and Italian versions.}
\label{tab:reliability_validity}
\end{threeparttable}
\end{table}

Moreover, we examined how the experimental manipulation of a robot's behavior transparency affected participants' perceptions. We used two-tailed t-tests and two-way ANOVAs to analyze the data, considering both the experimental manipulation and the participants' language (English, German, Italian) as factors. We initially planned to include participants' prior experience with robots as a covariate, but found that language significantly influenced this experience $(F(1, 895) = 5.32, p = .005, \eta_{\text{p}}^{2} = .01, 95\% \text{ CI } [0.00, 1.00])$, making it unsuitable as a covariate.
Our results (as shown in Table \ref{tab:CombinedANOVA}) revealed that participants perceived the robot's behavior as significantly more transparent in the high transparency condition compared to the low transparency condition $(t(874) = 29.88, p < .001, 95\% \text{ CI } [1.88, 2.14], d = 1.99)$. The ANOVA confirmed this main effect of the transparency manipulation. Additionally, we found a significant main effect of language on perceived transparency.
When comparing language groups, we found that German participants perceived higher transparency than English participants $(t(598) = 2.55, p = .031, 95\% \text{ CI } [0.07, 0.53], d = 0.21)$ (Post-hoc t-tests with Bonferroni correction). However, we didn't find significant differences between English and Italian participants $(t(598) = -2.18, p = .086, 95\% \text{ CI } [-0.48, -0.03], d = -0.18)$, or between German and Italian participants $(t(598) = 0.39, p = 1, 95\% \text{ CI } [-0.18, 0.27], d = 0.03)$. There was no significant interaction effect between transparency manipulation and language.

\begin{figure}[t]
    \centering
    \subfigure[The English version of the scale.]{
        \includegraphics[width=0.5\textwidth]{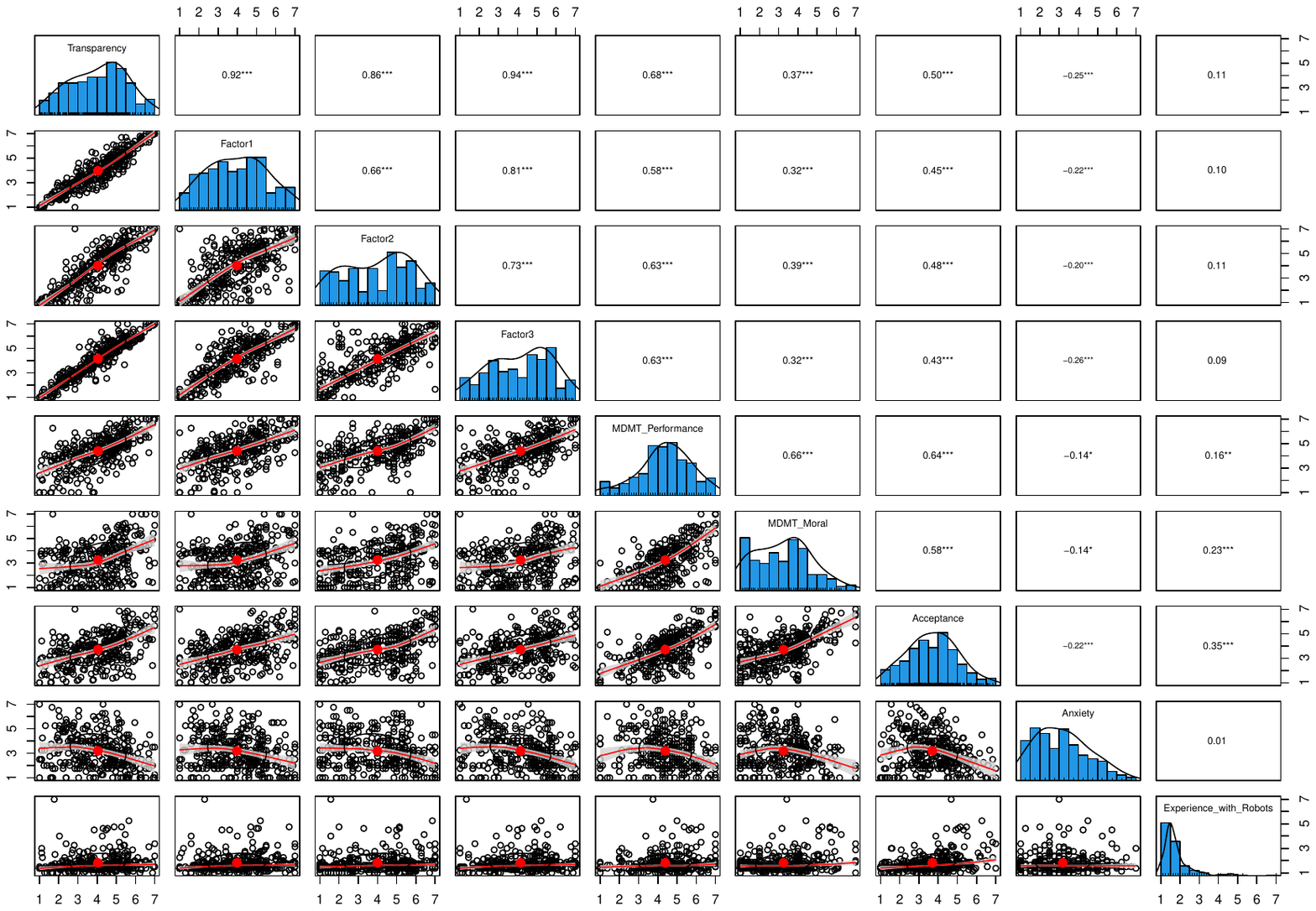}
        \label{correng2}
    }
    \vfill
    \subfigure[The German version of the scale.]{
        \includegraphics[width=0.48\textwidth]{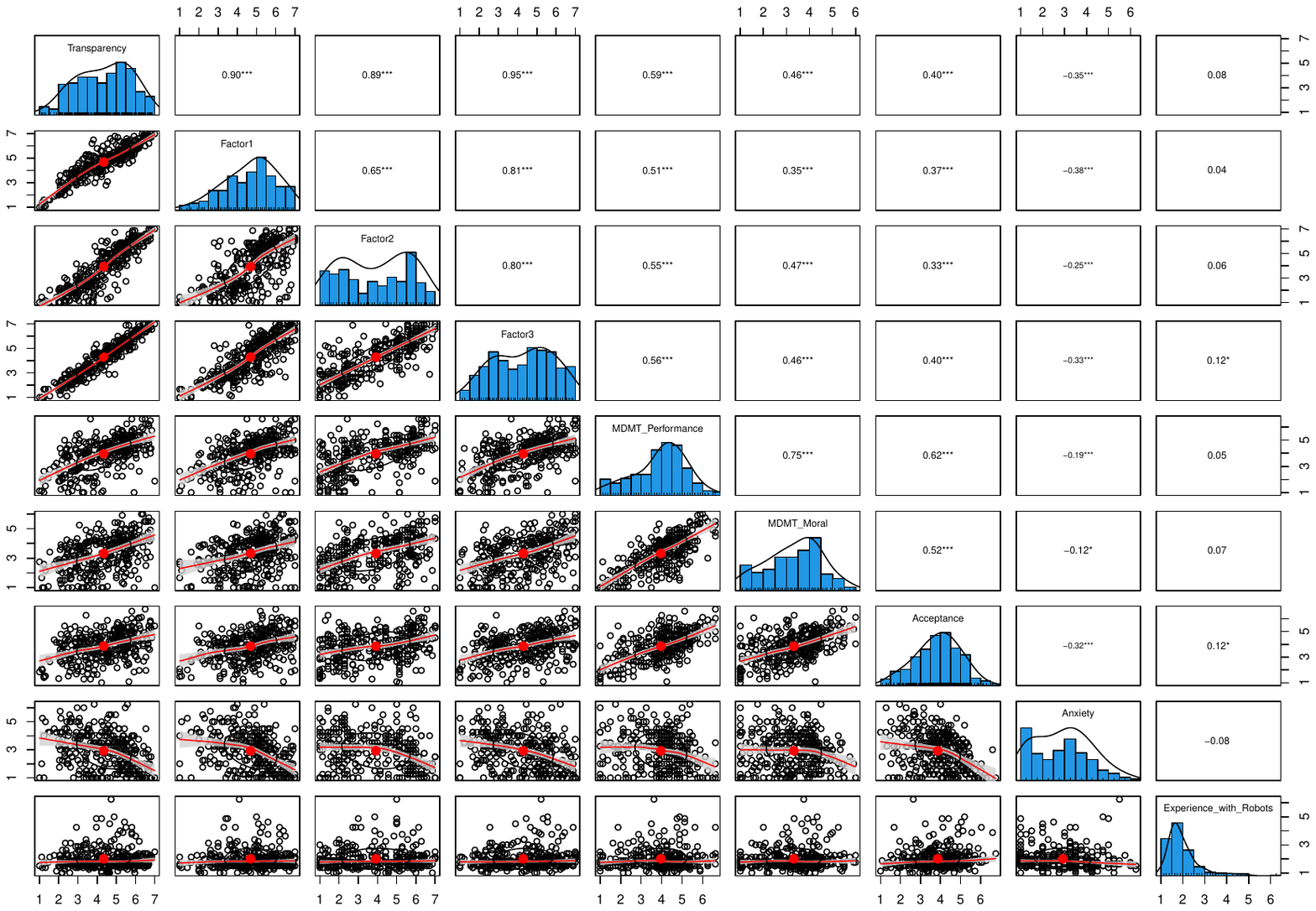}
        \label{corrger2}
    }
    \hfill
    \subfigure[The Italian version of the scale.]{
        \includegraphics[width=0.48\textwidth]{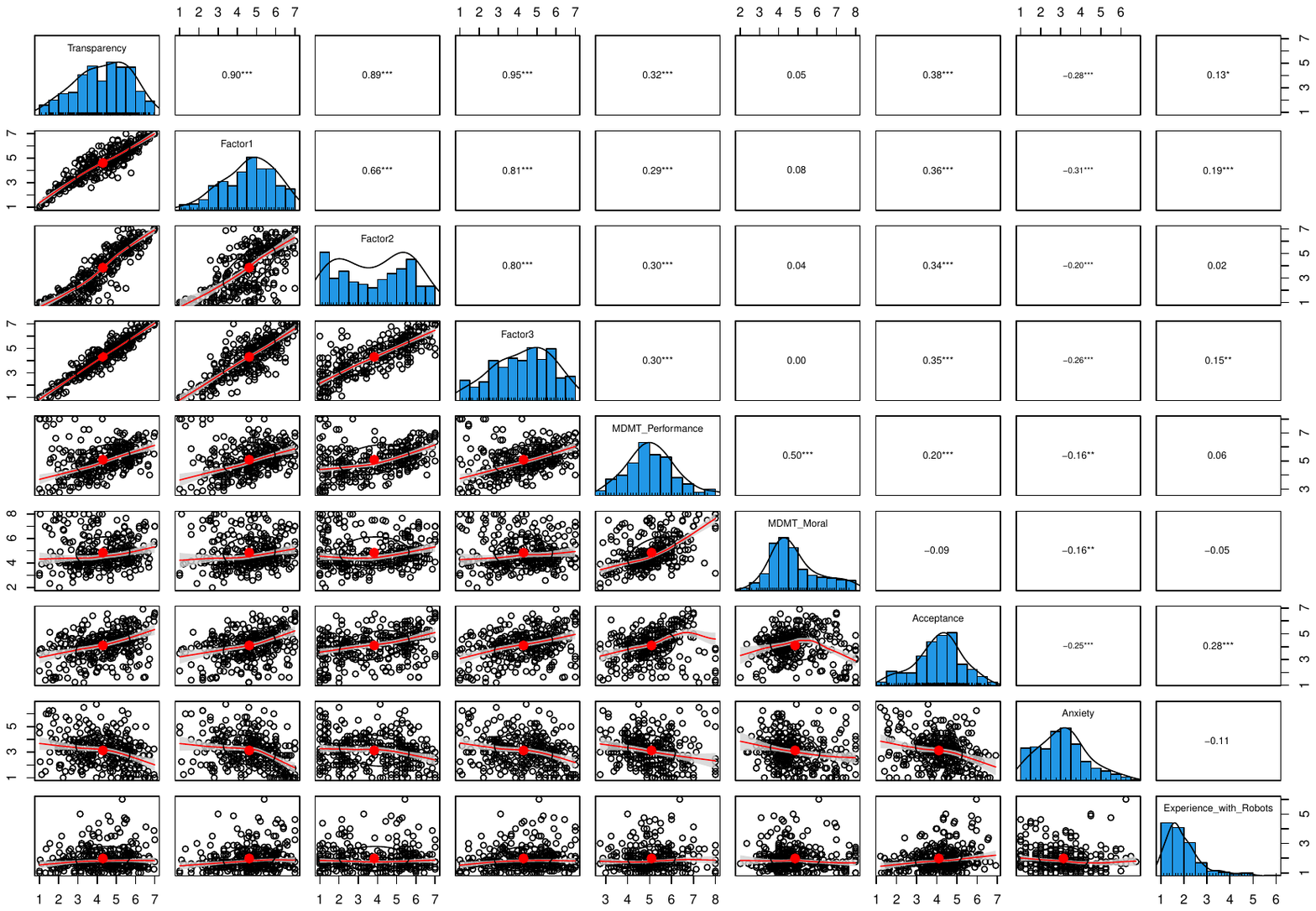}
        \label{corrita2}
    }
    \caption{Correlation matrices between the dependent and control variables of Experiment 2 across the three languages versions.}
    \label{correlation2}
\end{figure}

Furthermore, the three factors of the scale depending on the manipulation of transparency of the robot's behavior and across different language groups were investigated. For Factor 1, the high transparency condition showed significantly higher scores than the low transparency condition $(t(880) = 16.97, p < .001, 95\%$ $\text{ CI } [1.26, 1.59], d = 1.13)$. An ANOVA (see Table \ref{tab:CombinedANOVA}) showed that both the manipulated transparency and participants' language had significant main effects on Factor 1. Post-hoc tests showed that German participants scored lower than English participants $(t(590) = 6.00, p < .001, 95\% \text{ CI } [0.48, 0.94], d = 0.49)$, who scored lower than Italian participants $(t(590) = -5.46, p < .001, 95\% \text{ CI } [-0.87, -0.41], d = -0.45)$. However, there was no significant difference between German and Italian participants $(t(599) = 0.61, p = 1, 95\% \text{ CI } [-0.15, 0.28], d = 0.05)$, and no interaction effect was found. For Factor 2, the high transparency condition again scored significantly higher than the low transparency condition $(t(872) = 43.78, p < .001, 95\% \text{ CI } [2.79, 3.06], d = 2.91)$. An ANCOVA confirmed the main effect of manipulated transparency on Factor 2 but found no main effect of language. There was, however, a marginally significant interaction between the transparency manipulation and participant language. Lastly, Factor 3 also showed significantly higher scores in the high transparency condition compared to the low transparency condition $(t(880) = 24.36, p < .001, 95\% \text{ CI } [1.79, 2.11], d = 1.62)$. Finally, the ANOVA confirmed the main effect of the experimental manipulation on Factor 3 but found no significant effect of participant language and no interaction effect. Therefore, the transparency of robot behavior influences the scale, with some variations based on the language background of the participants.

Finally, how the experimental conditions and participants' language affected other dependent variables was examined. Performance trust was significantly higher in the high transparency condition $(t(857) = 11.06, p < .001, 95\% \text{ CI } [0.75, 1.08], d = 0.73)$. An ANOVA (as detailed in Table S21 of the Supplementary Information) showed that both the experimental manipulation and participants' language had significant effects on performance trust. Post-hoc tests (with Bonferroni correction) revealed that performance trust was significantly lower for the German participants than for the English participants, $t(594) = -3.99$, $p < .001$, $95\%$ CI $[-0.64, -0.22]$, $d = -0.33$, and for German participants than for Italian participants, $t(584) = -11.77$, $p < 001$, $95\%$ CI $[-1.32, -0.94]$, $d = -0.96$. Furthermore, performance trust was significantly lower for English participants than for Italian participants, $t(569) = -6.98$, $p < .001$, $95\%$ CI $[-0.90, -0.50]$, $d = -0.57$. There was also a significant interaction between experimental manipulation and language.
Moral trust was also higher in the high transparency condition $(t(885) = 5.67, p < .001, 95\% \text{ CI } [0.37, 0.75], d = 0.38)$. An ANCOVA (results presented in Table S22 of the Supplementary Information) revealed that the experimental manipulation and participants' language both had significant effects. Post-hoc tests (with Bonferroni correction) showed that Italian participants had significantly higher moral trust than both English participants,$t(590) = 14.33$, $p < .001$, $95\%$ CI $[1.38, 1.82]$, $d = 1.17$, and German participants, $t(596) = 15.12$, $p < 001$, $95\%$ CI $[1.33, 1.73]$, $d = 1.23$,  while there was no significant difference between English and German participants, $t(576) = 0.67$, $p = .501$, $95\%$ CI $[-0.14, -0.29]$, $d = 0.05$. An interaction effect was also observed.
Robot acceptance was higher in the high transparency condition $(t(884) = 6.53, p < .001, 95\% \text{ CI } [0.36, 0.67], d = 0.44)$. An ANCOVA (refer to Table S23 in the Supplementary Information) indicated that both experimental manipulation and language had significant effects. Post-hoc tests (with Bonferroni correction) showed that Italian participants had the highest acceptance, significantly higher than both English, $t(590) = 3.96$, $p < .001 $, $95\%$ CI $[0.20, 0.60]$, $d = 0.32$, and German participants, $t(599) = 2.53$, $p = .048$, $95\%$ CI $[0.06, 0.42]$, $d = 0.21$. There was no significant difference between German and English participants, $t(586) = 1.63$, $p = .289$, $95\%$ CI $[-0.03, 0.36]$, $d = 0.13$. No interaction effect was identified by the ANOVA.

Anxiety towards the robot was lower in the high transparency condition $(t(897) = -4.47, p < .001, 95\% \text{ CI } [-0.56, -0.22], d = -0.30)$. An ANCOVA (as illustrated in Table S24 of the Supplementary Information) showed that both experimental manipulation and language had significant effects. Post-hoc tests (with Bonferroni correction) revealed that German participants had significantly less anxiety than English participants, $t(593) = -2.42$, $p = .039$, $95\%$ CI $[-0.48, -0.05]$, $d = -0.20$. No significant differences were found between English and Italian participants, $t(588) = 0.49$, $p = 1 $, $95\%$ CI $[-0.16, 0.26]$, $d = 0.04$, or between German and Italian participants, $t(597) = -2.08$, $p = .138$, $95\%$ CI $[-0.42, -0.01]$, $d = -0.17$. There was no interaction effect.

Considering the above findings, they suggest that our proposed scale for measuring perceived transparency correlates well with established measures of HRI. The consistent pattern of higher transparency corresponding with increased trust and acceptance and decreased anxiety supports the validity of our scale.

\begin{table}[h!]
\centering

\begin{tabular}{llccccccc}
\hline
\multicolumn{2}{c}{Parameter}     & df  & SS     & MS     & $F$ value & $p$ value            & $\eta p^2$   & 95\% CI         \\ \hline
\multicolumn{9}{c}{\textbf{Overall Score of the Transparency}} \\ \hline
\multicolumn{2}{l}{Transparency condition}            & 1   & 910.30 & 910.30 & 907.36  & \textless .001 *** & .50   & {[}0.47, 1.00{]} \\
\multicolumn{2}{l}{Language}                          & 2   & 19.60  & 9.80   & 9.76    & \textless .001***  & .02   & {[}0.01, 1.00{]} \\
\multicolumn{2}{l}{Transparency condition: Language} & 2   & 3.60   & 1.80   & 1.80    & .167              & $<.01$ & {[}0.00, 1.00{]} \\
\multicolumn{2}{l}{Residuals}                         & 895 & 897.9  & 1.0    &         &                    &       &                  \\ \hline
\multicolumn{9}{c}{\textbf{Factor 1}} \\ \hline
\multicolumn{2}{l}{Transparency condition}            & 1   & 456.40  & 456.40 & 306.72  & \textless .001 *** & .26 & {[}0.22, 1.00{]} \\
\multicolumn{2}{l}{Language}                          & 2   & 98.00   & 49.00  & 32.92   & \textless .001***  & .07 & {[}0.04, 1.00{]} \\
\multicolumn{2}{l}{Transparency condition: Language} & 2   & 0.80    & 0.40   & 0.276   & .759               & $<.01$ & {[}0.00, 1.00{]} \\
\multicolumn{2}{l}{Residuals}                         & 895 & 1331.90 & 1.50   &         &                    &     &                  \\ \hline
\multicolumn{9}{c}{\textbf{Factor 2}} \\ \hline
\multicolumn{2}{l}{Transparency condition}            & 1   & 1925.70 & 1925.70 & 1921.13 & \textless .001 *** & .68 & {[}0.66, 1.00{]} \\
\multicolumn{2}{l}{Language}                          & 2   & 4.60    & 2.30    & 2.28    & .103               & .01 & {[}0.00, 1.00{]} \\
\multicolumn{2}{l}{Transparency condition: Language} & 2   & 5.90    & 3.0     & 2.94    & .053               & .01 & {[}0.00, 1.00{]} \\
\multicolumn{2}{l}{Residuals}                         & 895 & 897.10  & 1.00    &         &                    &     &                  \\ \hline
\multicolumn{9}{c}{\textbf{Factor 3}} \\ \hline
\multicolumn{2}{l}{Transparency condition}            & 1   & 858.00  & 858.00 & 594.44  & \textless .001 *** & .40 & {[}0.36, 1.00{]} \\
\multicolumn{2}{l}{Language}                          & 2   & 6.30    & 3.10   & 2.18    & .114               & .01 & {[}0.00, 1.00{]} \\
\multicolumn{2}{l}{Transparency condition: Language} & 2   & 6.80    & 3.40   & 2.36    & .095               & .01 & {[}0.00, 1.00{]} \\
\multicolumn{2}{l}{Residuals}                         & 895 & 1291.90 & 1.40   &         &                    &     &                  \\ \hline
\end{tabular}
\begin{tablenotes}
\item Note: df = Degrees of freedom; SS = Sum of Squares; MS = Mean Squares;\\ *p< .05; **p< .01; ***p< 0.001;
\end{tablenotes}
\caption{Results of the 2-way ANOVA with the manipulation of the transparency and the language of participants as independent variables}
\label{tab:CombinedANOVA}

\end{table}

%% file: 5_discussion.tex

The aim of the present research was to develop and validate a scale to assess the perceived transparency of a robot's behavior. Experiment 1 showed that the perceived transparency of a robot consists of three factors. Building upon these findings, we propose the Transparency Of RObots Scale (TOROS), based on a three-factor model: The first factor, \textit{Illegibility} comprises items expressing difficulty in comprehending the robot's functioning, objectives, and processes (e.g., ``The robot's overall functioning is a mystery to me'', ``I cannot comprehend the robot's inner processes''). The second factor, \textit{Explainability}, is based on items evaluating perceived quality, clarity, and usefulness of the robot's explanations about its actions and states (e.g., ``The robot explains complex tasks in a way that is easy to understand'', ``The robot provides clear explanations for its actions''). The third factor, \textit{Predictability} represents items assessing the users' ability to anticipate or foresee the robot's future actions based on its current behavior (e.g., ``It is easy for me to foresee the robot's future actions'', ``The robot's behavior is predictable''). While conceptually related  \cite{angelopoulos2023robot,endsley2017here,alonso2018system}, the three proposed TOROS factors offer a new perspective on measuring the perceived robot transparency, supported by empirical data from three countries. In addition, the results showed that these factors were all sensitive to the experimental manipulation of the explainability, legibility, and predictability of a robot's behavior. Experiment 2 confirmed the factorial structure of the scale in three languages: English, German, and Italian. 

Additionally, in both Experiments 1 and 2, the TOROS scale demonstrates a good convergent validity with factors related to transparency, namely trust towards robots and acceptance of robots \cite{schor2024mind,fischer2018increasing,aquilino2024trust}. The experimental manipulation of transparency in both Experiments 1 and 2 had an effect on trust and acceptance and, therefore, confirms that the transparency of a robot's behavior is an important determinant of trust and acceptance in HRI. Hence, TOROS has the potential to provide more accurate estimations of the influence of perceived transparency on trust, acceptance, and other constructs that transparency is supposed to determine.

Taken together, our results confirm that TOROS represents a reliable and valid measure of the perceived transparency of a robot's behavior. Besides, it underlies the discrepancy between theorizing about and implementation of transparency in HRI, and how individuals perceive it. More specifically, the results suggest that any manipulation of transparency mostly influences \textit{Explainability}. Interestingly, we found no interaction effect between manipulated explainability, legibility, and predictability of the robot's behavior on the factors of the scale in Experiment 1, yet all these factors had main effects on the different subscales of the TOROS. This suggests that distinguishing explainability, legibility, and predictability of a robot from perceived transparency of a robot's behaviors is important. Existing theories of transparency and the factorial structure of TOROS are consistent in terms of what constitutes transparency. Nevertheless, the way transparency is implemented in a robot does not result in equivalent perceptions of transparency in a user's mind (e.g., Making a robot more predictable does not only result in higher perceived predictability). This remains to be confirmed in other scenarios. Future research should delve more into the psychological mechanisms that determine the perceived transparency of a robot or any artificial agent. 

Interestingly, in Experiment 2, despite TOROS being administered after participants saw an entire video with a robot reaching its goal, an effect of the experimental manipulation of transparency on \textit{Predictability} factor was still detected. This goes against what can be referred to as the \textit{Valley of the normal}: People tend to find ordinary events to be retrospectively predictable \cite{kahneman_noise_2021}. This is due to the fact that understanding processes of resolved events are ``backward-looking'': When confronted with unexpected but also unsurprising events, people tend to examine the past in a causal thinking process. This induces them to conclude that such events are self-explanatory and to overestimate their predictability. This phenomenon is known as the hindsight bias \cite{fischhoff_hindsight_1975, roese_hindsight_2012}, and could have skewed the answers of participants. Indeed, as perceived transparency was assessed after a robot's behavior was fully resolved, a ceiling effect for \textit{Predictability} could have been observed, and yet was not. Further examination of this phenomenon in future research is required.

Despite the promising results of the TOROS scale, the present research does not come without methodological limitations: For instance, a crucial limitation pertains to the measurement invariance results across the different language versions of the scale. Even though configural and metric invariance were achieved, TOROS did not demonstrate full scalar and residual invariance.  Besides, there were slight differences in perceived transparency across languages, particularly regarding English and German participants, as well as between English and Italian participants. However, it is important to note that achieving only partial measurement invariance is not uncommon in cross-cultural research, especially when dealing with complex psychological constructs like transparency in HRI. As \cite{robitzsch2023full} argue, partial invariance can still allow for meaningful cross-group comparisons. The strong internal consistency and convergent validity demonstrated across all three language versions suggest that the scale is reliable and valid within each language context. Additionally, the observed differences between languages in terms of perceived transparency were small, and almost no interaction between the language of the participant and the manipulation of transparency was observed.
Moreover, whereas our approach using image vignettes and videos of real scenarios provided a strong foundation for scale development, we recognize the potential value of extending the validation to real-life HRI. This step, however, should be viewed as an avenue for future research rather than a limitation. Our current methodology aligns with established practices in scale development within HRI \cite{boateng_best_2018, schrum_concerning_2023}, as demonstrated by seminal works like \cite{bartneck2009measurement,carpinella2017robotic}. The controlled nature of our approach allows for precise manipulation of transparency levels, which is crucial for initial scale validation.

%% file: 6_conclusions.tex
As robots become more sophisticated and prevalent in our society, the need to measure how transparent they are to humans becomes increasingly important. Yet, until now, a standardized method to measure this critical aspect of robot behavior has been lacking. This work addresses this gap by developing and validating the first comprehensive scale to assess the perceived transparency of robotic systems, termed TOROS. Through a rigorous three-stage process involving 1,223 participants, we have created a robust tool encompassing 26 items and comprised of three factors: \textit{Illegibility}, \textit{Explainability}, and \textit{Predictability}. The scale demonstrates high cross-linguistic reliability and validity across English, German, and Italian languages. We believe that the proposed scale can serve as a valuable tool in various HRI experiments to examine the effects of transparency-related aspects on other phenomena. For instance, it can be employed in studies of human-robot collaboration to evaluate how transparency impacts team performance, in learning and adaptation research to track changes in the understanding of the robot as users gain experience with it, and in error recovery and management to assess the effectiveness of error handling strategies. The scale can also be used to investigate how different robot designs and behaviors influence perceived transparency and how this, in turn, affects the overall interaction quality.

Future research should focus on translating the scale into more languages and examining its performance in real HRI scenarios or in interaction with different artificial agents. Follow-up works can use the scale to understand the determinants of perceived transparency in HRI and contribute to a better understanding of the psychological mechanisms at play. This tool opens up new avenues for research and has the potential to significantly enhance our understanding of transparency in robotics, ultimately leading to the development of more effective and user-friendly robotic systems.